\documentclass{article}
\usepackage{spconf,amsmath,graphicx}

\usepackage{epsfig}
\usepackage{setspace}
\usepackage{mathtools, cuted}
\usepackage{lipsum, color}

\usepackage{amsmath}
\usepackage{amssymb,amsthm}
\usepackage{booktabs} 
\usepackage{bm,algorithm,algpseudocode,threeparttable,subfigure,rotating,multirow,array,graphics,textcomp}
\usepackage{graphicx,gensymb}

\newcommand{\xqedhere}[2]{%
	\rlap{\hbox to#1{\hfil\llap{\ensuremath{#2}}}}}

\usepackage{enumitem}
\usepackage{textcomp,upgreek}

\graphicspath{{Figures/}}

\usepackage{caption}
\let\OLDthebibliography\thebibliography
\renewcommand\thebibliography[1]{
	\OLDthebibliography{#1}
	\setlength{\parskip}{0pt}
	\setlength{\itemsep}{0pt plus 0.3ex}
}

\title{Expression Conditional GAN for 
	\\ Facial Expression-to-Expression Translation}
\name{Hao Tang$^1$, Wei Wang$^2$, Songsong Wu$^3$, Xinya Chen$^4$, Dan Xu$^5$, Nicu Sebe$^1$, Yan Yan$^6$
}
\address{$^1$University of Trento \, $^2$EPFL \, $^3$Nanjing University of Posts and Telecommunications \\ 
    $^4$Huazhong University of Science and Technology \, $^5$University of Oxford \, $^6$Texas State University
}
\begin{document}
%

\maketitle
\begin{abstract}
	
	In this paper, we focus on the facial expression translation task and propose a novel Expression Conditional GAN  (ECGAN) which can learn the mapping from one image domain to another one based on an additional expression attribute. 
	The proposed ECGAN is a generic framework and is applicable to different expression generation tasks where specific facial expression can be easily controlled by the conditional attribute label.
	Besides, we introduce a novel face mask loss to reduce the influence of background changing.
	Moreover, we propose an entire framework for facial expression generation and recognition in the wild, which consists of two modules, i.e., generation and recognition.
	Finally, we evaluate our framework on several public face datasets in which the subjects have different races, illumination, occlusion, pose, color, content and background conditions.
	Even though these datasets are very diverse, both the qualitative and quantitative results demonstrate that our approach is able to generate facial expressions accurately and robustly.
	
\end{abstract}


\vspace{-0.3cm}
\section{Introduction}
\vspace{-0.3cm}

Recently, Generative Adversarial Networks (GANs) have shown to capture complex image data with numerous applications in computer vision and image processing.
For example, Pix2pix \cite{isola2016image} can translate an image  from one domain to another one in a supervised way, i.e., the training image pairs are required. 
However, obtaining paired training data can be difficult and expensive in some cases as indicated in~\cite{zhu2017unpaired}.
To tackle this limitation, Zhu et al.~propose CycleGAN~\cite{zhu2017unpaired}, in which the model can learn the mapping function from one  domain to another one with unpaired training data.
Similar ideas have been proposed in~\cite{liu2017unsupervised,taigman2016unsupervised,kim2017learning,yi2017dualgan}. 
Despite these efforts, facial expression translation remains a challenging task due to the fact that the expression changes are non-linear~\cite{tang2019attention,pumarola2018ganimation}. 

To overcome the aforementioned challenging, we propose a novel Expression Conditional GAN (ECGAN) for facial expression translation based on CycleGAN~\cite{zhu2017unpaired}. 
ECGAN can generate faces with different emotions which are conditioned on the input expression attribute vector. 
Our work is inspired by IcGAN \cite{perarnau2016invertible} which factorizes an input image into a latent representation and conditional information using the trained encoders. 
By changing the conditional information, the generator network combines the same latent representation and the changed conditional information to generate an image that satisfies the changed encoded constraints. 

\begin{figure}[!t] \small
	\centering
	\includegraphics[width=1\linewidth]{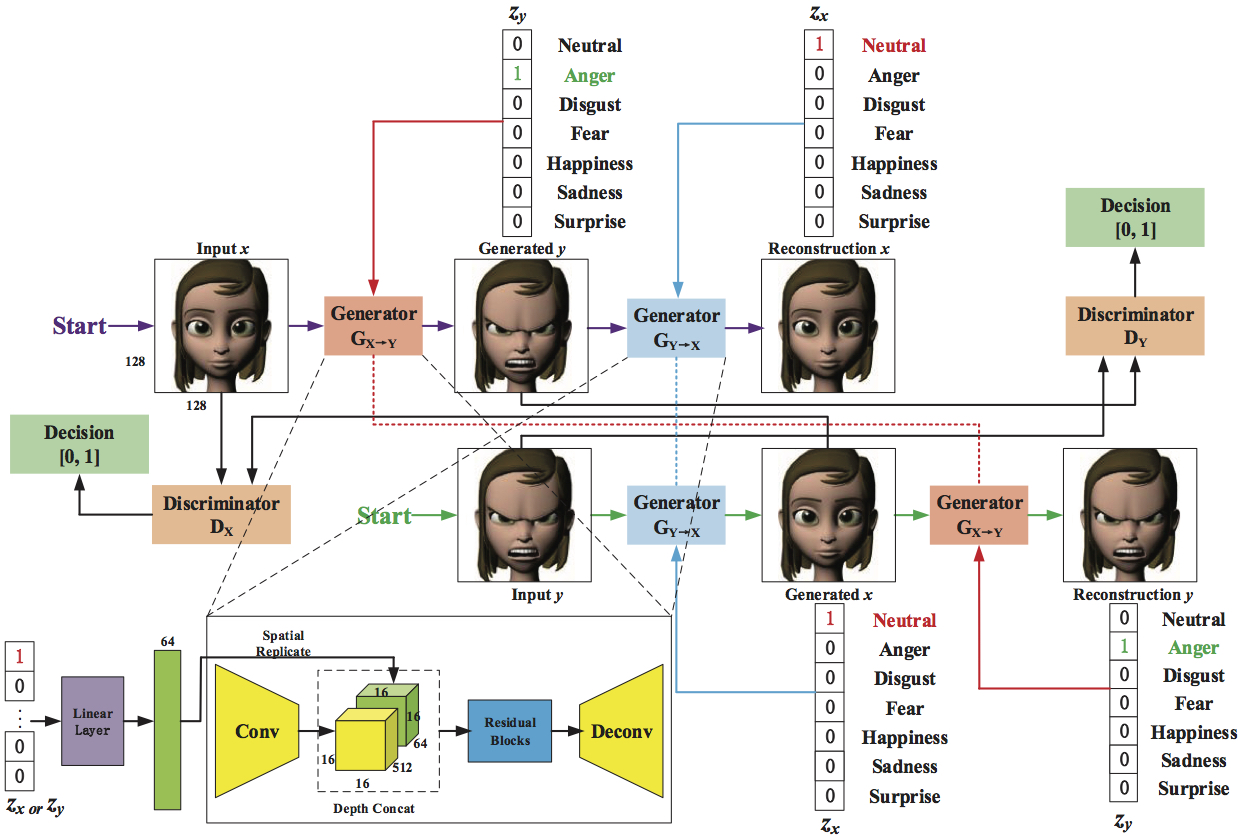}
	\caption{Framework of the proposed ECGAN. 
		Image $X$ can be converted to a new modified expression face image $Y$ guided by the facial expression attribute vector $z$. The expression attribute vector $z$ is concatenated with the image representation in the convolution layers.}
	\label{fig:excyclegan}
	\vspace{-0.6cm}
\end{figure}

In this paper, we present another strategy in which the conditional attribute vector is concatenated with the image representation in the convolutional layers, as shown in Fig.~\ref{fig:excyclegan}. 
The conditional attribute is represented by a vector, which is used to distinguish each attribute from the others. 
In the attribute vector, only the element which corresponds to the label is set to 1 while the rest of them are set to 0. 
Then the vector is concatenated with the image embedding vector at the bottleneck which is a fully connected layer of generator $G_{X{\rightarrow}Y}$. 
In the generator $G_{Y{\rightarrow}X}$, we change the expression vector by swapping the corresponding two expressions.
The conditional label can be used to guide the transformation from one expression to another one. 
For instance, as shown in Fig.~\ref{fig:excyclegan}, the anger label corresponds to an angry face with an open mouth. 
A correspondence between the anger label and the open mouth is built. 
During training time, GAN can learn this correspondence automatically.
Thus, ECGAN can reshape a face (e.g., mouth) by adding the conditional vector.

Moreover, we introduce a novel face mask loss to reduce the influence of background changing similar to \cite{liang2017generative}.
We also present a complete framework for facial expression translation and recognition in the wild.
Our framework comprises of two modules, i.e.,  translation and recognition. 
ECGAN allows us how to map a face with neutral expression to the faces with other expressions (e.g., anger, disgust), and vice versa.
We can explicitly control the expression of a face image via the conditional expression vector, which can be potentially useful in several applications, such as data augmentation and facial expression profiling. 
Then, we rely on a face recognition model to evaluate the generated images of ECGAN.
Overall, our contribution is three-fold:
(1) We propose ECGAN, which allows us to generate and modify real images of faces conditioned on arbitrary facial expressions.
(2) We propose a novel face mask loss for alleviating the influence of background changing.
(3) We  propose a new VGG score to evaluate the generated images by GAN models. 

\vspace{-0.3cm}
\section{Related Work}
\vspace{-0.3cm}

\sloppypar \textbf{Generative Adversarial Networks (GANs)}~\cite{goodfellow2014generative}  have achieved impressive performance on image generation tasks~\cite{brock2018large,isola2016image,zhu2017unpaired,tang2019attention,wang2018high,tang2018gesturegan}.
Moreover, conditional GANs~\cite{mirza2014conditional} are proposed to generate meaningful images that meet a certain requirement, where the conditioned label is employed to guide the image generation process. 
The conditioned labels can be discrete class labels \cite{perarnau2016invertible}, text descriptions \cite{mansimov2015generating,reed2016generative}, object keypoints \cite{reed2016learning}, human skeleton \cite{tang2018gesturegan}, semantic maps \cite{mo2018instagan,tang2019multi} or reference images~\cite{isola2016image,zhu2017unpaired}. 
The conditional models with images have tackled a lot of problems, e.g., image editing \cite{perarnau2016invertible}, text-to-image translation~\cite{mansimov2015generating,zhang2017stackgan}, image-to-image translation~\cite{isola2016image} and video-to-video translation~\cite{wang2018video,bansal2018recycle}. 

\noindent \textbf{Image-to-Image Translation} learns a mapping function between different image domains using CNNs. 
Pix2pix~\cite{isola2016image} employs a conditional GAN to learn a mapping function from input to output images in a supervised way. 
Wang et al.~\cite{wang2018high} further propose Pix2pixHD model, which can turn semantic label maps into photo-realistic images or synthesizing portraits from face label maps.
Similar ideas have also been applied to many other tasks, such as \cite{sangkloy2016scribbler,tang2018gesturegan,zhu2017toward,tang2019multi}. 
However, these methods need to use the paired input-output data for training, which is not feasible for some applications.
To overcome this limitation, 
Zhu et al.~\cite{zhu2017unpaired}  propose CycleGAN, which learns the mappings between two different image domains with the unpaired data.
Moreover, many other GAN variants are proposed to tackle the unpaired image-to-image translation task, such as ~\cite{liu2017unsupervised,taigman2016unsupervised,choi2017stargan,benaim2017one,tang2019dual,kim2017learning,yi2017dualgan,gokaslan2018improving,mejjati2018unsupervised,lu2017conditional,pumarola2018ganimation,chang2018pairedcyclegan,tang2019attention}.

\noindent \textbf{Face Editing.} Face analysis has a wide range of applications, such as face completion \cite{li2017generative}, 
hair modeling \cite{chai2015high}, aging \cite{antipov2017face}, image-to-sketch translation \cite{chen2018sketchygan,tang2019attribute}.
For example, Taigman et al.~\cite{taigman2016unsupervised} propose  Domain Transfer Network (DTN) for  face-to-emoji translation task.
Several other works \cite{perarnau2016invertible,lu2017conditional,larsen2015autoencoding}  focus on human face attributes (e.g., bald, bangs, black hair, blond hair, eyeglasses, heavy makeup, male, mustache, pale skin) translation. 
For instance, 
Larsen et al.~\cite{larsen2015autoencoding} use a combination of Variational Autoencoder (VAE) and GAN to generate face samples with visual attribute vectors added to their latent representations. 
Shu et al.~\cite{shu2017neural} present an end-to-end GAN that infers a face-specific disentangled representation of intrinsic face properties, including shape (i.e., normals), albedo, lighting, and an alpha matte.
In this work, we focus on the arbitrary facial expression translation task with unpaired training data.

\vspace{-0.3cm}
\section{Formulation}
\vspace{-0.3cm}

GANs~\cite{goodfellow2014generative} are composed of two competing modules, i.e., a generator $G_{X\rightarrow Y}$ and a discriminator $D_Y$ (Where $X$ and $Y$ denote two different domains), which are iteratively trained competing against with each other in the manner of two-player minimax.
CycleGAN~\cite{zhu2017unpaired} includes two mappings $G_{X\rightarrow Y}{:} X {\rightarrow}Y$ and $G_{Y\rightarrow X}{:} Y{\rightarrow}X$, and two adversarial discriminators $D_X$ and $D_Y$.
The generator $G_{X\rightarrow Y}$ maps $X$ from the source domain to the target domain $Y$ and tries to fool the discriminator $D_Y$, whilst the $D_Y$ focuses on improving itself in order to be able to tell whether a sample is a generated sample or a real data sample.
The similar to the generator $G_{Y\rightarrow X}$ and the discriminator $D_X$.
More formally, let $x_i{\in} X$ and $y_j{\in} Y$ (For simplicity, we usually omit the subscript $i$ and $j$.) denote the training images in source and target image domain, respectively.
We intent to learn a mapping function between $X$ domain and $Y$ domains with training data $\{x_i\}_{i=1}^N$ and $\{y_j\}_{j=1}^M$.

\subsection{Objective Function}
Our ECGAN objective contain several losses, we will introduce each of them, respectively.

\noindent \textbf{Adversarial loss.} 
We  apply a least square loss \cite{mao2017least} to stabilize our model during training.
The least square loss is more stable than the negative log likelihood objective and more faster than Wasserstein GAN (WGAN) \cite{arjovsky2017wasserstein} to converge:
\begin{equation}\footnotesize
\begin{aligned}
& \mathcal{L}_{lsgan}(G_{X \rightarrow Y}, D_Y, X, Y) =  \mathbb{E}_{y\sim{p_{\rm data}(y)}}[(D_Y(y)-1)^2] \\
& +  \mathbb{E}_{x\sim{p_{\rm data}}(x), z\sim{p_z(z)}}[D_Y( G_{X \rightarrow Y}(x,z))^2]，
\end{aligned}
\label{equ:legan}
\end{equation}
where $G_{X \rightarrow Y}$ tries to generate images $G_{X \rightarrow Y}(x, z)$ that look similar to images from domain $Y$, while $D_Y$ aims to distinguish between translated samples $G_{X \rightarrow Y}(x, z)$ and real samples $y$.
$G_{X \rightarrow Y}$ aims to minimize this objective against an adversary $D_Y$ that tries to maximize it.
We have a similar  loss for generator $G_{Y \rightarrow X}$ and discriminator $D_X$ as well.

\noindent \textbf{Cycle Consistency Loss.}
Note that CycleGAN \cite{zhu2017unpaired} is different from Pix2pix \cite{isola2016image} in the way that the training data in CycleGAN is unpaired. 
CycleGAN introduces the cycle consistency loss to enforce forward-backward consistency.
The cycle consistency loss can be regarded as ``pseudo'' pairs of training data 
even though we do not have the corresponding data in the target domain which corresponds to the input data from the source domain.
To include facial expression conditional constraint $z$ as part of the input to the generator and discriminator of  CycleGAN, the loss of cycle consistency is reformulated as follows:
\begin{equation}\footnotesize
\begin{aligned}
& \mathcal{L}_{cyc}(G_{X \rightarrow Y}, G_{Y \rightarrow X})  \\ 
= & \mathbb{E}_{x\sim{p_{\rm data}}(x), z\sim{p_z}(z)}[\Arrowvert G_{Y \rightarrow X}(G_{X \rightarrow Y}(x, z))-x\Arrowvert_1] \\
+ & \mathbb{E}_{y\sim{p_{\rm data}}(y), z\sim{p_z}(z)}[\Arrowvert G_{X \rightarrow Y}(G_{Y \rightarrow X}(y, z))-y\Arrowvert_1].
\end{aligned}
\end{equation}

\begin{figure}[!t]
	\centering
	\includegraphics[width=1\linewidth]{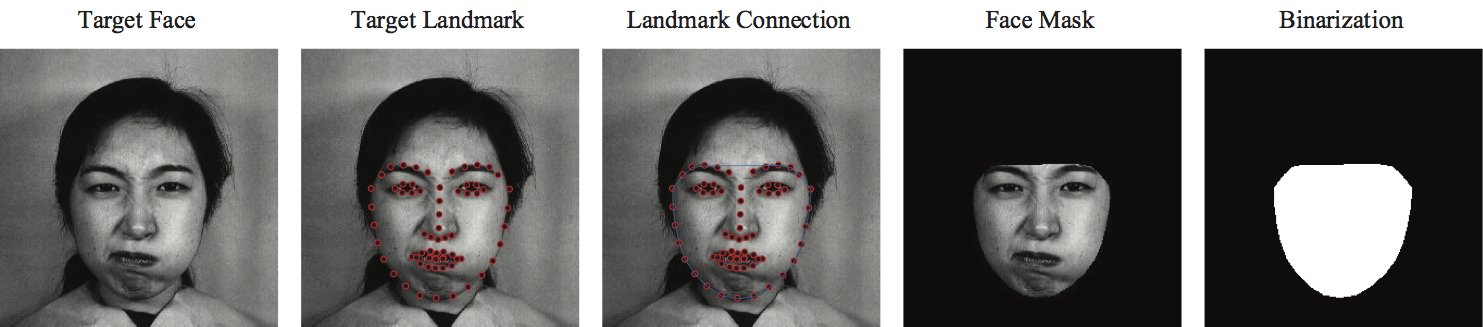}
	\caption{Process of computing the face mask.}
	\label{fig:mask_loss}
	\vspace{-0.6cm}
\end{figure}

\noindent \textbf{Context Loss.}
The pixel-wise MSE loss is used as a context loss in \cite{dong2016image,shi2016real}.
However, since the pixel-wise MSE loss often lacks high-frequency content which results in perceptually unsatisfying solutions with overly smooth textures.
Ledig et al.~\cite{ledig2016photo} introduces the VGG loss, which is closer to perceptual similarity.
The formulation of the VGG loss as follows:
\begin{equation} \footnotesize
\begin{aligned}
& \mathcal{L}_{content}^{VGG_{Y \rightarrow X}} {=} \frac{1}{W_{i,j}H_{i,j}}\sum_{w=1}^{W_{i,j}}\sum_{h=1}^{H_{i,j}}(\phi_{i,j}(X)_{w,h}-\phi_{i,j}G_{Y \rightarrow X}(y)_{w,h})^2,
\end{aligned}
\end{equation}
where, $\phi_{i,j}$ indicate the feature map obtained by the $j$-th convolution before the $i$-th max-pooling layer within VGG net~\cite{simonyan2014very}, $W_{i,j}$ and $H_{i,j}$ are the dimensions of the respective feature maps within the VGG network.
Therefore, the final loss $\mathcal{L}_{content}^{VGG} = \mathcal{L}_{content}^{VGG_{Y \rightarrow X}}{+} \mathcal{L}_{content}^{VGG_{X \rightarrow Y}}$.

\noindent \textbf{Identity Preserving Loss.}
To reinforce the identity of the face while converting, a face identity preserving loss \cite{taigman2016unsupervised} is adopted to preserve the identity.
\begin{equation}\footnotesize
\begin{aligned}
& \mathcal{L}_{identity}(G_{X\rightarrow Y}, G_{Y\rightarrow X})= \mathbb{E}_{x\sim{p_{\rm data}}(x)}[\Arrowvert G_{Y\rightarrow X}(x)-x\Arrowvert_1] + \\
& \mathbb{E}_{y\sim{p_{\rm data}}(y)}[\Arrowvert G_{X\rightarrow Y}(y)-y\Arrowvert_1]
\end{aligned}
\end{equation}
In such way, generators will take into consideration the identity problem through the back-propagation of the identity loss.

\noindent \textbf{Face Mask Loss.}
In order to eliminate the influence brought by background changes, we propose a novel loss that add a face mask $M$ to the $L_1$ loss such that the face is given larger weight than the background, as shown in Fig.~\ref{fig:mask_loss}.
We apply OpenFace \cite{amos2016openface} to extract face landmark.
The formulation of face mask is given as follows with $\odot$ as the pixel-wise multiplication:
\begin{equation} \footnotesize
\begin{aligned}
\mathcal{L}_{mask}^{Y\rightarrow X}= \Arrowvert (G_{Y{\rightarrow}X}(G_{X\rightarrow Y}(x \odot M_x)) - x \odot M_x)\Arrowvert_1,
\end{aligned}
\end{equation}  
face mask $M_x$ are set to 1 for foreground and 0 for background and applying a set of morphological operations such that it is able to approximately cover the whole face.
Thus, $\mathcal{L}_{mask}=\mathcal{L}_{mask}^{Y\rightarrow X}+\mathcal{L}_{mask}^{X\rightarrow Y}$.

\noindent \textbf{Full Objective.}
Consequently, the complete objective loss is:
\begin{equation} \footnotesize
\begin{aligned}
\mathcal{L}(G_{X \rightarrow Y}, G_{Y \rightarrow X}, D_X, D_Y) =  &\mathcal{L}_{cGAN}+  \lambda_1 \mathcal{L}_{cyc}(G_{X \rightarrow Y}, G_{Y \rightarrow X}) + \\
& \lambda_2 \mathcal{L}_{context} +  \lambda_3 \mathcal{L}_{identity} + \lambda_4 \mathcal{L}_{mask},
\end{aligned}
\label{eqn:allloss}
\end{equation}
where $\lambda_1$, $\lambda_2$, $\lambda_3$ and $\lambda_4$ are parameters controlling the relative relation of objectives terms.

\begin{figure}[!t] \footnotesize
	\centering
	\includegraphics[width=1\linewidth]{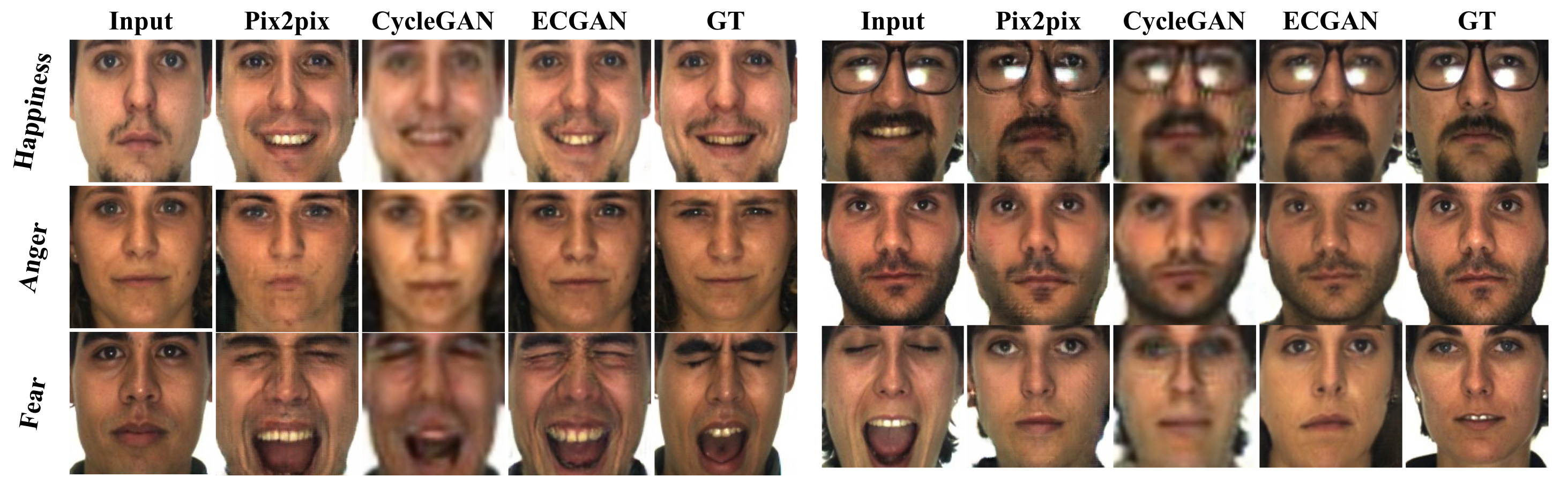}
	\caption{Different methods for facial expression generation (Left) and neutralization (Right). Form left to right: input, Pix2pix trained on paired data, CycleGAN, ECGAN and Ground Truth (GT). Note that images are cropped for visualization.}
	\label{fig:comparison_expression}
	\vspace{-0.6cm}
\end{figure}

\vspace{-0.3cm}
\section{Experiments}
\vspace{-0.3cm}

In this section, we first introduce the details of the employed datasets in our experiments, then we demonstrate the results and discussions of generation and recognition steps respectively.

\noindent \textbf{Datasets.}
We employ several datasets to validate our model. These datasets contains faces with different races and they have different illumination, occlusion, pose conditions and backgrounds.  See the supplementary materials for details.

\noindent \textbf{Setup.}
We use the same training setups as CycleGAN \cite{zhu2017unpaired}.
Adam optimizer~\cite{kingma2014adam} with a batch size of 1 is used.
The initial learning rate for Adam optimizer is 0.0002 and $\beta_1$ of Adam is 0.5. 
For fair comparisons, all models were trained for 200 epochs. 
Training and testing stages are conducted out on an Nvidia TITAN Xp GPU with 
12GB memory. 

\noindent \textbf{Competing Models.}
We employ state-of-the-art image translation models, i.e., CycleGAN \cite{zhu2017unpaired}, Pix2pix \cite{isola2016image} as our baselines. 
Note that Pix2pix \cite{isola2016image} is trained on paired data.
For a fair comparison, we implement both baselines using the same setups as our approach. 

\noindent \textbf{Evaluation Metrics.}
We provide both  qualitative and quantitative results.
Qualitatively, the images generated by different methods as shown in Fig.~\ref{fig:comparison_expression}.
Quantitatively, the expression recognition accuracy score is employed to evaluate whether the generated images wear the correct expressions.
To this end, we propose a novel VGG Score which is similar to ``FCN Score'' in \cite{isola2016image} and Inception Score \cite{salimans2016improved} as the score of accuracy.
The definition of the Inception Score is $exp(E_x[KL(p(y|x) || p(y))])$, where $x$ is an image, $p(y|x)$ is the inferred class label probabilities given $x$ by the pre-trained Inception network and $p(y)$ is the marginal distribution over all images.
The VGG score is defined as $E_x(p(y|x, z))$, where $z$ is the conditioning label.
Overall, the differences between VGG score and Inception Score \cite{salimans2016improved} are as follows.
First, the VGG Score is calculated using a pre-trained VGG network, while the Inception Score is calculated using a pre-trained Inception network. 
Second, even though both the VGG Score and the Inception Score are defined to maximize inferred probabilities in order to guarantee that the generated images are meaningful,
the difference is that the Inception Score includes an extra term to maximize the entropy of the marginal distributions to encourage the diversity of the generated images. 

\begin{figure}[!t]\scriptsize
	\centering
	\setcounter{subfigure}{0}
	\subfigure[AR dataset.]      {\includegraphics[height=0.14\linewidth]{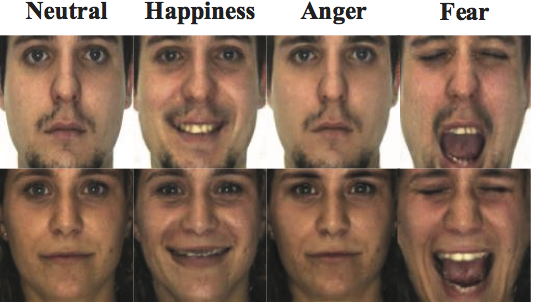}}
	\subfigure[Yale dataset.]    {\includegraphics[height=0.14\linewidth]{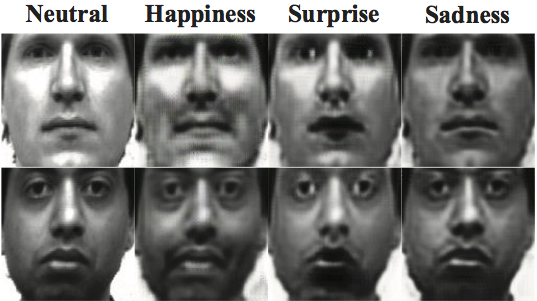}}
	\subfigure[JAFFE dataset.] {\includegraphics[height=0.14\linewidth]{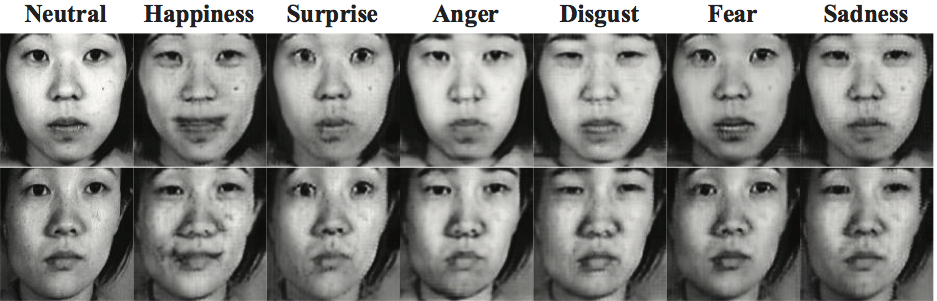}}
	\subfigure[FERG dataset.]  {\includegraphics[height=0.155\linewidth]{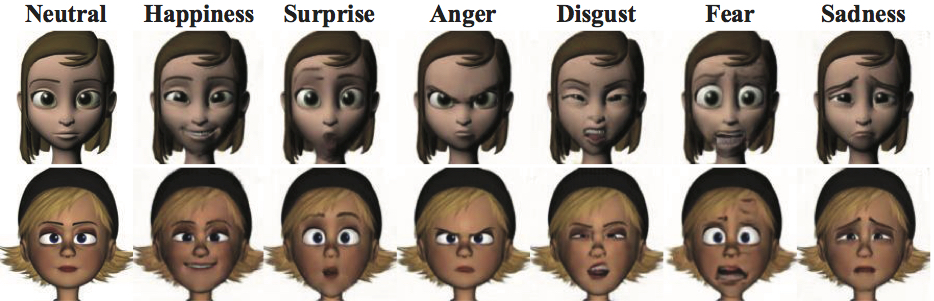}} 
	\subfigure[3DFE dataset.]  {\includegraphics[height=0.155\linewidth]{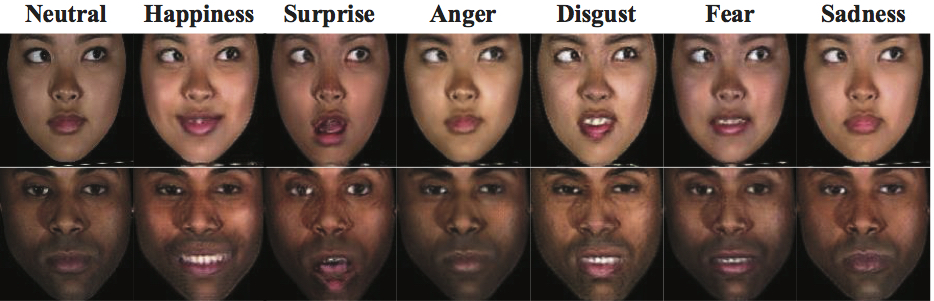}}
	\caption{Example results of  ECGAN. The neutral expression is the input and the others expressions are the output. We can observe that even though the subjects in all datasets with significant differences, our method consistently generates high-quality images, which shows our method is very insensitive to changing skin color, posture, illumination or occlusion.}
	\label{fig:result}
	\vspace{-0.4cm}
\end{figure}

\noindent \textbf{Qualitative Evaluation.}
Fig.~\ref{fig:comparison_expression} demonstrates the images generated by our method and the baselines. 
We can observe that the results generated by CycleGAN tend to be more blurry compared with Pix2pix and ECGAN. 
ECGAN adopts the proposed face mask loss to guide the generators to focus on the face regions.
Though Pix2pix also generates images with competitive quality, the model can only be trained with paired data. 
In contrast, our ECGAN produces good quality images without the requirement of paired data. 
To exhaustively validate the superiority of our ECGAN, Fig.~\ref{fig:result} provides more generation results.
We can see that ECGAN generalizes well to the unseen data.
We also observe that even though the subjects in all datasets have different races, poses, skin colors, illumination conditions and occlusions, our method consistently generates high-quality images. 
This demonstrates that our method is very robust.

\begin{table}[!t]\small
	\centering
	\caption{AMT Score of different methods.}
	\begin{tabular}{lccc} \hline
		Method                                          & CycleGAN \cite{zhu2017unpaired} & Pix2pix \cite{isola2016image}     & ECGAN (Ours) \\ \hline
		AMT Score                                     & 11.68   & \textbf{40.37}  & 35.32 \\ \hline
	\end{tabular}
	\label{tab:amt}
	\vspace{-0.4cm}
\end{table}

\begin{table}[!t]\small
	\centering
	\caption{VGG Score (\%) of different methods.}
	\begin{tabular}{lccc} \hline
		Method                                         & Train Set  & Test Set    & VGG Score \\ \hline
		baseline                                        & original   & original       & 74.77 \\
		CycleGAN \cite{zhu2017unpaired} & +generated & original   & 76.41 \\
		CycleGAN \cite{zhu2017unpaired} & original   & generated  & 77.78 \\ 
		Pix2pix \cite{isola2016image}        & +generated & original   & 82.63 \\	
		Pix2pix \cite{isola2016image}        & original   & generated  & \textbf{83.24} \\ \hline	
		ECGAN (Ours)                              & +generated & original   & 78.13 \\
		ECGAN (Ours)                              & original   & generated  & 80.32 \\ \hline
	\end{tabular}
	\label{tab:score}
	\vspace{-0.6cm}
\end{table}

\begin{figure}[!t] \scriptsize
	\centering
	\setcounter{subfigure}{0}
	\subfigure[VGG score.]{\includegraphics[width=0.4\linewidth]{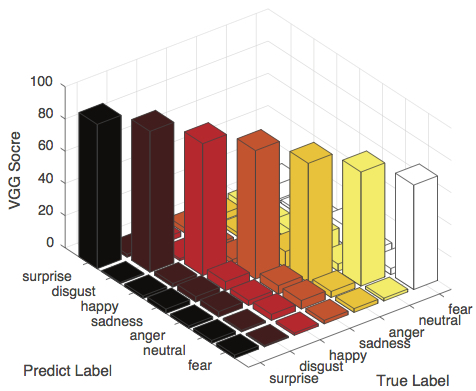}}
	\subfigure[Expression space.] {\includegraphics[width=1.3in, height=1in]{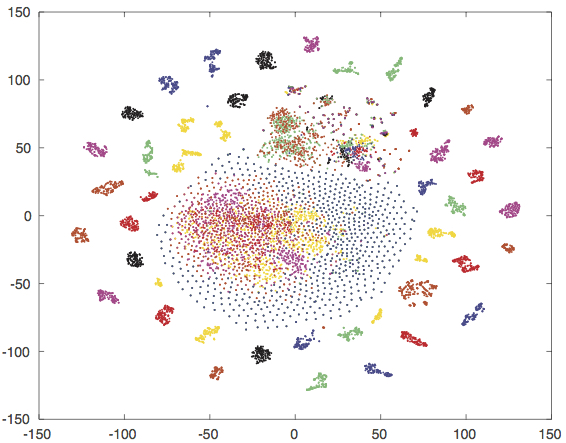}}\\
	\subfigure[Legend.]{\includegraphics[width=0.16\linewidth, height=.6in]{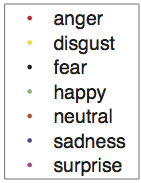}} 
	\subfigure[Loss and gradient.]{\includegraphics[width=.6\linewidth, height=.6in]{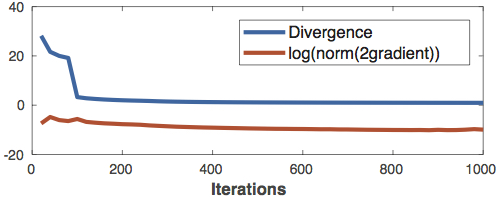}} 
	\caption{(a) VGG score. 
		(b-d) Feature space of the generated facial expressions. Each color represents a expression.}
	\label{fig:tsne}
	\vspace{-0.6cm}
\end{figure}

\noindent \textbf{Quantitative Evaluation.}
We follow \cite{zhu2017unpaired} to conduct the ``real vs fake'' perceptual studies on Amazon Mechanical Turk (AMT) to assess the realism of the generated images.
Results are shown in Table~\ref{tab:amt}.
We can see that the proposed method is significantly better than CycleGAN, but a little worse than Pix2pix.
Moreover, we employ the expression recognition accuracy to evaluate the correctness of the generated expressions.
The intuition is that if the generated images are realistic, then (i) the classifiers trained on both the real images and the generated images will be able to boost the accuracy of the real images (in this situation, the generated images work as augmented data.) (ii) the classifiers trained on real images will also be able to classify the synthesized image correctly.
VGG~\cite{simonyan2014very} is adopted as deep feature extractor for our facial expression recognition task. 
Detailed recognition performance is reported in Table~\ref{tab:score},
when the model is trained only with the original training data, and tested on the testing data,
the score is 74.77\%. When the generated images are added to the training set as augmented data, the recognition score of the testing data is increased to 78.13\%.
Besides, to validate that our model can generate the correct expressions, we replaced the test set by the generated images and achieve 80.32\% recognition rate (Fig.~\ref{fig:tsne}(a)), which demonstrates the effectiveness of our method since the generative images have a slight better performance than the ground truth images. 
Moreover, we also conduct the experiment of expression clustering to visulize the distribution of the generated images, t-SNE \cite{maaten2008visualizing} is adopted to visualize the 4,096-D deep feature on a two dimensional space.
Fig.~\ref{fig:tsne}(b-d) illustrates the deep feature space of the generated images as well as the evolving of the loss and gradient in the training stage. 
Note that the generated images with the same expressions are classified into the same clusters according to their representations in the deep feature space, which reveals that our ECGAN method can generate images with correct expressions. 

\vspace{-0.3cm}
\section{Conclusion}
\vspace{-0.3cm}

We propose Expression Conditional GAN (ECGAN) for the facial expression generation task. 
The main technical contribution is the proposed Conditional CycleGAN which utilizes the expression label to guide the facial expression generation process.
In ECGAN, the adversarial loss is modified to include a conditional expression feature vectors as parts of the inputs to the generator and discriminator networks. 
The expression attribute vector is utilized to represent the expression label.
Experimental results demonstrate that our method not only presents compelling results but also achieves competitive results on facial expression recognition task.

\noindent \textbf{Acknowledgment:}
We acknowledge the gift donation from Cisco, Inc for this research.


\footnotesize
\bibliographystyle{IEEEbib}
\bibliography{strings}

\end{document}